\documentclass[letterpaper, 10 pt, conference]{ieeeconf}  

\IEEEoverridecommandlockouts                              

\title{\LARGE \bf
Aggressive Trajectory Generation for A Swarm of Autonomous Racing Drones
}
\usepackage{graphicx}
\usepackage{subfigure}
\usepackage{amsmath}
\usepackage{mathtools}
\usepackage{float}
\usepackage{amsfonts,amssymb}

\author{Yuyang Shen$^{1}$, Jinming Xu$^{1}$, Jin Zhou$^{1}$, Danzhe Xu$^{2}$, Fangguo Zhao$^{3}$, Jiming Chen$^{1}$, and Shuo Li$^{1}$
\thanks{$^{1}$Authors are with the College of Control Science and Engineering, Zhejiang University, Hangzhou 310027, China
        {\tt\small shuo.li@zju.edu.cn}
        }%
\thanks{$^{2}$Danzhe Xu is with the Department of Automation, Zhejiang University of Technology, Hangzhou 310023,China.
        }%
\thanks{$^{3}$Fangguo Zhao is with the
School of Automation, Northwestern Polytechnical University, Xi’an 710072, China }%
}

\begin{document}

\maketitle
\thispagestyle{empty}
\pagestyle{empty}


\begin{abstract}

Autonomous drone racing is becoming an excellent platform to challenge quadrotors' autonomy techniques including planning, navigation and control technologies. However, most research on this topic mainly focuses on single drone scenarios. In this paper, we describe a novel time-optimal trajectory generation method for generating time-optimal trajectories for a swarm of quadrotors to fly through pre-defined waypoints with their maximum maneuverability without collision. We verify the method in the Gazebo simulations where a swarm of $5$ quadrotors can fly through a complex $6$-waypoint racing track in a $35m \times 35m$ space with a top speed of $14m/s$. Flight tests are performed on two quadrotors passing through $3$ waypoints in a $4m \times 2m$ flight arena to demonstrate the feasibility of the proposed method in the real world. Both simulations and real-world flight tests show that the proposed method can generate the optimal aggressive trajectories for a swarm of autonomous racing drones. The method can also be easily transferred to other types of robot swarms.  

\end{abstract}

\section{INTRODUCTION}

Autonomous drone racing is an excellent platform to challenge drones' autonomous aggressive flight techniques including environment perception, trajectory planning, state estimation and control, etc. The rules for this racing are quite simple that the drones have to fly fully autonomously and pass through the gates in a certain sequence and the fastest one wins the racing. It has to be admitted that although the speed of autonomous racing drones has increased significantly, there are still large gaps between professional human racing pilots and the best autonomous drone racing techniques. One of these challenges is the time-optimal trajectory generation for one single racing drone or a swarm of racing drones in complex racing tracks.

\begin{figure}[hbt]
    \centering
    \includegraphics[scale=0.04, clip]{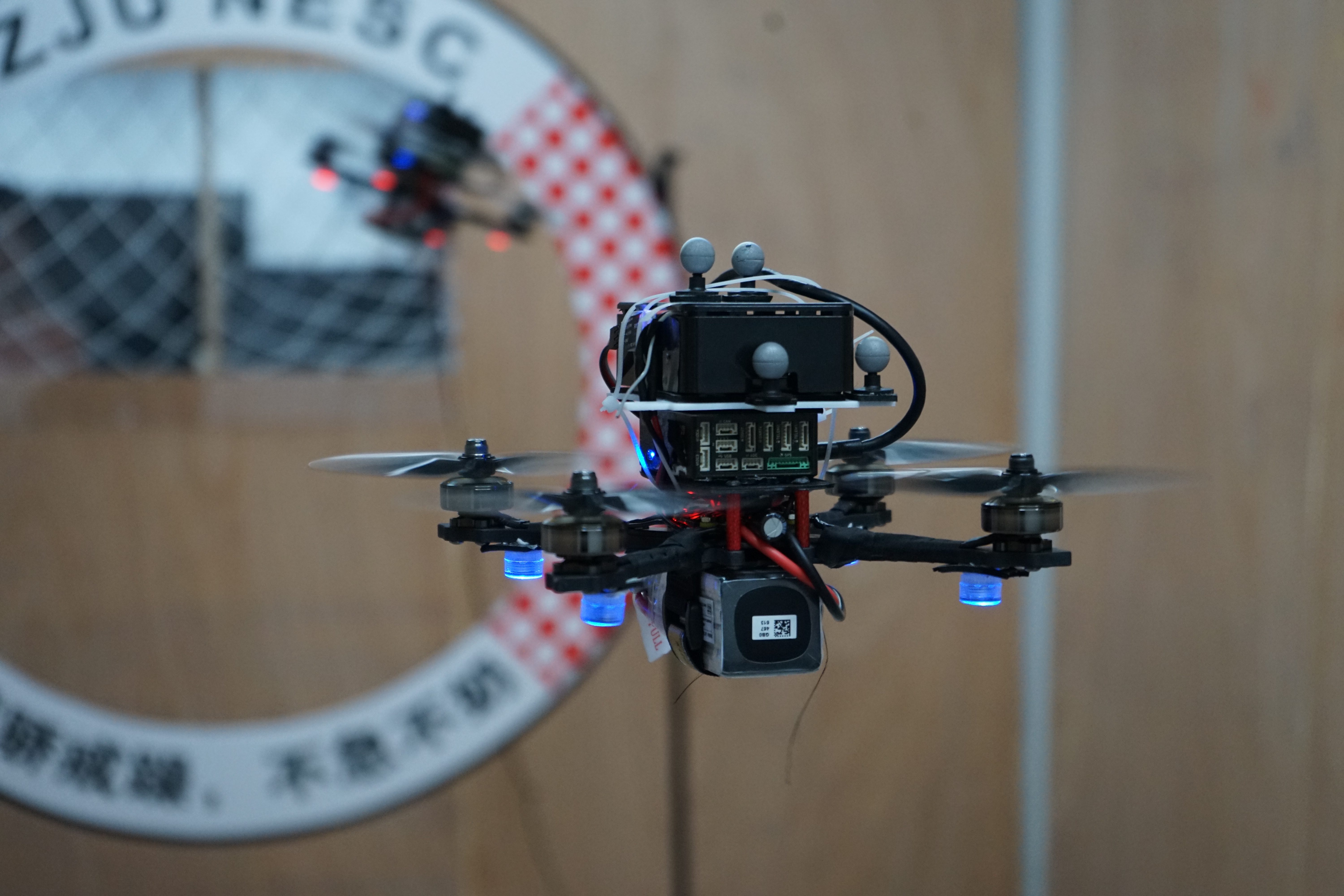}
    \caption{The flying platforms used in the experiment fly through the waypoints.}
    \label{fig:two drone racing photo}
\end{figure}

Before 2016, there was no research conducted on autonomous drone racing but aggressive trajectory generation and control techniques had been well developed, among which differential flatness theory-based methods have shown their strengths in generating agile trajectories for quadrotors passing through pre-defined waypoints \cite{mellinger2011minimum,faessler2017differential}. Until now, these methods are still one of the most commonly used quadrotor trajectory generation methods. Although these methods can generate agile trajectories, due to their polynomial character, the generated trajectories cannot be time optimal which makes the racing drones not able to fly at their extreme. One milestone for autonomous drone racing is the IROS 2016 autonomous drone racing which was designed to provide a chance for research teams and hobbyist teams to communicate and compete their newest techniques and finally beat human pilots \cite{moon2017iros}. From this point, the IROS autonomous drone racing became an annual event until 2019. At this early stage, the speed of the autonomous racing drone gradually increased year by year, but the speed was still far from human pilots. For example, the winners' speeds were around $0.6m/s$ in 2016 \cite{jung2018direct} and $0.7m/s$ in 2017 \cite{moon2019challenges}. The MAVLab, TU Delft made a Bebop quadrotor fly through a five-gate racing track in a complex basement environment with a top speed of $1.7m/s$ in 2018 \cite{li2020autonomous}. And later, they developed a 72-gram autonomous racing drone that can fly a simple four-gate track with a top speed of $2.6m/s$ \cite{li2020visual}. The winner of IROS 2019 autonomous drone racing, the Robotics and Perception Group (RPG) from the University of Zurich, has pushed the flying speed to $2.5m/s$ \cite{kaufmann2019beauty}. One important turning point was the 2019 Artificial Intelligence Robotic Racing Competition, also known as AlphaPilot Challenge, where the winner, the MAVLab from TU Delft, achieved the top speed of $9.2m/s$ \cite{de2022sensing} while the second place, the RPG, achieved the top speed of $8m/s$ \cite{foehn2022alphapilot}. Unfortunately, according to the report, the speed of the autonomous racing drone was very close to the human pilots but failed to surpass them. Another milestone of autonomous drone racing is Scaramuzza's work in 2021, where they developed a novel trajectory generation method called CPC that could generate truly time-optimal trajectories and guide the quadrotor to fly at the highest speed of $19m/s$ \cite{foehn2021time}. Their method finally outperformed human expert drone pilots in a drone-racing task. 


From the competitions and research mentioned above, there was only one single drone flying through the racing track. If we have a look at the first-person-view (FPV) drone racing, human pilots can compete against up to 5 opponents simultaneously \cite{hanover2023autonomous}. Thus, how to have multiple autonomous racing drones compete simultaneously is a very interesting and challenging topic in the robotics community. To the best of the authors' knowledge, currently, there is no research on generating time-optimal trajectories for a swarm of autonomous racing drones. Currently, the related research work mainly focuses on quadrotor swarms like formation flight. Most of them used optimized polynomials to guide the quadrotors while avoiding each other which cannot  guarantee the optimality of the generated trajectories \cite{kushleyev2013towards,honig2018trajectory}. Others used decentralized methods for some complex tasks. For example, in \cite{mcGuire2019minimal}, a swarm of micro quadrotors were used for the search and rescue tasks in an unknown environment and Duisterhof et al. used a swarm of quadrotors to seek gas-leaking in cluttered environments \cite{duisterhof2021sniffy}. Zhou et al. realized a swarm of quadrotors flying through a bamboo forest at 
 the speed of $3m/s$ \cite{zhou2022swarm}. However, 'time-optimal' is not the target of their solutions and the quadrotors' flying speed is still far from their flight envelopes' boundaries.


 Hence, in this paper, we 

 \begin{enumerate}
            \item extend the CPC to a swarm of autonomous racing drones to make the quadrotors fly through racing tracks fully autonomously with their extreme maneuverability and arrive at the goal with the minimum flying time.
	    \item reformulate the optimization problem for a swarm of racing drones to make constraining collision avoidance available in generating time-optimal aggressive trajectories.  
	    \item valid and analyze our approach in simulation where $5$ quadrotors fly through $6$ waypoints with a maximum speed of $14m/s$ and also in the real-world flight tests to demostrate the feasibility of the proposed method.
\end{enumerate}

\section{METHODOLOGY}
\subsection{The quadrotors' model}
In this paper, we adopt the same quadrotors' dynamics model as the one presented in \cite{foehn2021time}. For the reader's convenience, we list the dynamics model here. It should be noted that we add the left subscript $i$ to the variables/states to denote that they belong to the $i^{th}$ quadrotor. 

\begin{align}
\prescript{}{i}{\dot{\mathbf{x}}} = \mathbf{f}_{dyn}(\prescript{}{i}{\mathbf{x}},\prescript{}{i}{\mathbf{u}}) =
\begin{cases}
\prescript{}{i}{\mathbf{v}} \\
\mathbf{g}+\frac{1}{\prescript{}{i}{m}}\mathbf{R}(\prescript{}{i}{\mathbf{q}}){\prescript{}{i}{\mathbf{T}}} \\
\frac{1}{2}\Lambda(\prescript{}{i}{\mathbf{q}})
\begin{bmatrix}
0 \\ \prescript{}{i}{\boldsymbol{\omega}} 
\end{bmatrix} \\
 \prescript{}{i}{\mathbf{J}}^{-1}( \prescript{}{i}{\boldsymbol{\tau}} - \prescript{}{i}{\boldsymbol{\omega}}\times \prescript{}{i}{\mathbf{J}}\prescript{}{i}{\boldsymbol{\omega}})
\end{cases}
\label{equ:simulation model}
\end{align}

where 

\begin{align*}
    {\prescript{}{i}{\mathbf{T}}} = \begin{bmatrix}
        0 \\  0 \\ \sum \prescript{}{i}{{T}_s}
    \end{bmatrix}
\end{align*}

is the thrust vector of the quadrotor and 

\begin{align*}
    {\prescript{}{i}{\boldsymbol{\tau}}} = \begin{bmatrix}
        {\prescript{}{i}{l}}/ \sqrt{2}( \prescript{}{i}{{T}_1} + \prescript{}{i}{{T}_2} -\prescript{}{i}{{T}_3} -\prescript{}{i}{{T}_4}) \\
       {\prescript{}{i}{l}}/ \sqrt{2}( -\prescript{}{i}{{T}_1} + \prescript{}{i}{{T}_2} +\prescript{}{i}{{T}_3} -\prescript{}{i}{{T}_4}) \\
       {\prescript{}{i}{c}}_{\tau}( \prescript{}{i}{{T}_1} - \prescript{}{i}{{T}_2} +\prescript{}{i}{{T}_3} -\prescript{}{i}{{T}_4})
    \end{bmatrix}
\end{align*}

is the torque vector of the quadrotor. In the equations above $\prescript{}{i}{\mathbf{v}}$, $\prescript{}{i}{\mathbf{q}}$ and $\prescript{}{i}{\boldsymbol{\omega}}$ are velocity, quaternions and angular velocity of the $i^{th}$ quadrotor, respectively. $\mathbf{R}(\prescript{}{i}{\mathbf{q}})$ is the rotation matrix. $\prescript{}{i}{l}$ is the arm length. $\prescript{}{i}{{T}_s}$ are the thrust of the rotors, which are the inputs of the dynamic system (\ref{equ:simulation model}). 

\begin{figure}[!h]
    \centering
    \vspace{0.5cm}
     \subfigure[During the optimization process, the arrival time $t_{end}$ decreases and the time between each node decreases as a result.]{\includegraphics[scale=0.4,trim={6cm 0.8cm 28cm 6cm},clip]{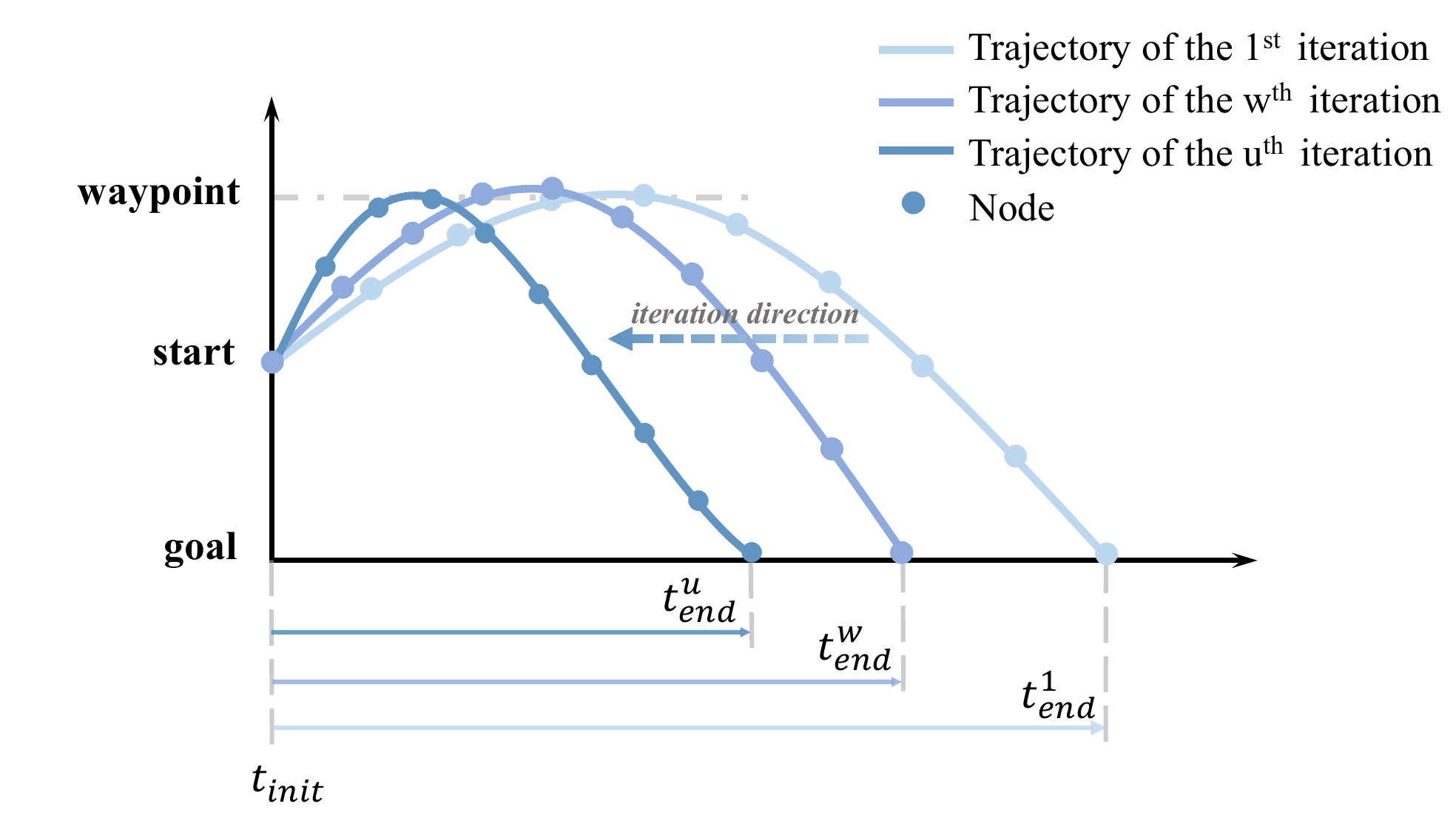}} 
     \subfigure[In terms of multi-drone racing scenarios, different arrival time leads to the mismatch of the time of corresponding nodes.]{\includegraphics[scale=0.4,trim={27.2cm 3.2cm 7cm 7cm},clip]{Figures/origin2.pdf}} 
    \caption{Illustrative sketch of the CPC and its deficiency in multi-drone racing scenarios.}
    \label{fig:foehn sketch} 
\end{figure}

\begin{figure*}[htb]
    \centering
    \subfigure[]{\includegraphics[scale=0.35,trim={0 1.3cm 30cm 0}, clip]{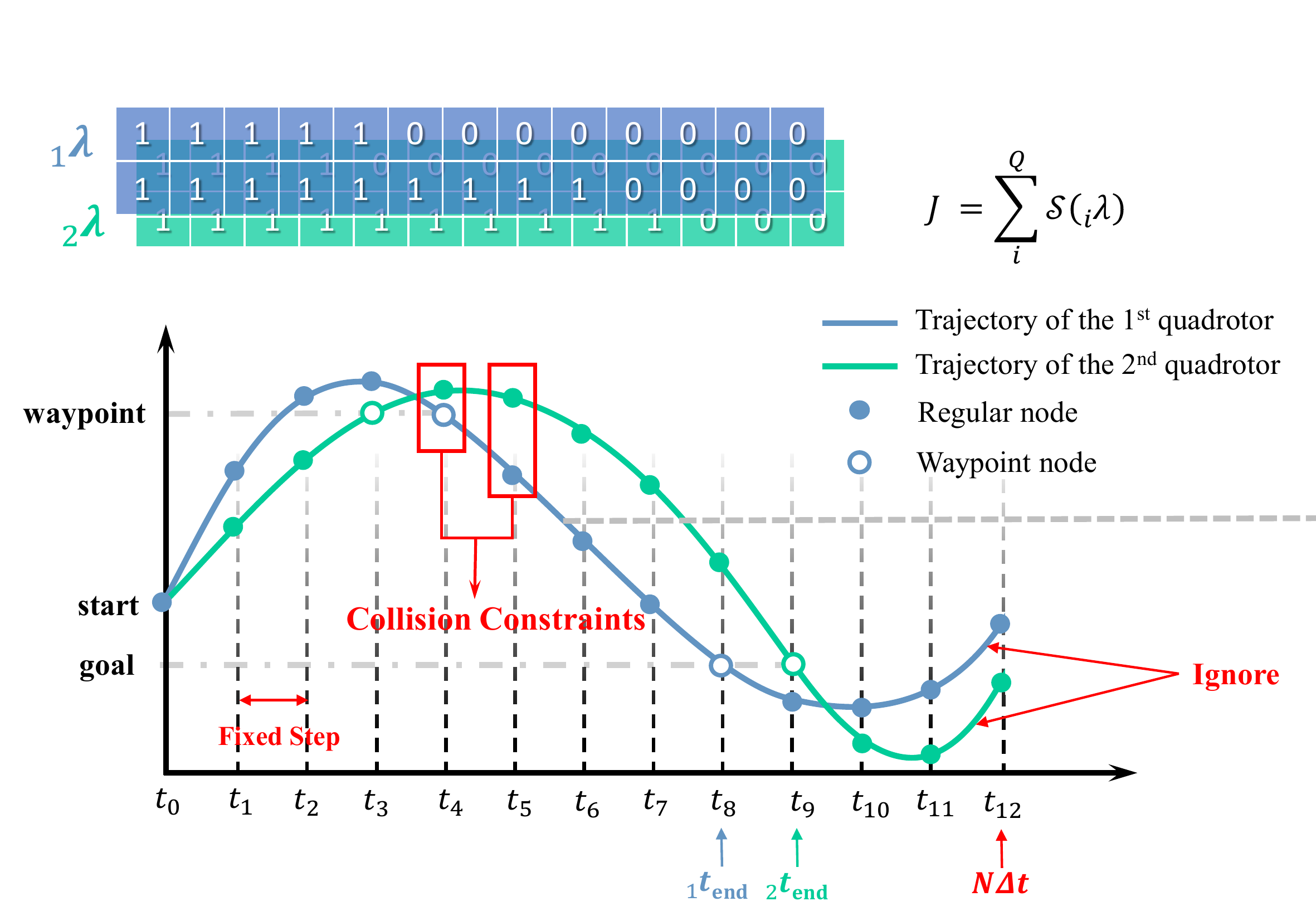}} 
    \subfigure[]{\includegraphics[scale=0.35,trim={28cm 1.3cm 0cm 0}, clip]{Figures/algorithm02.pdf}} 
    \caption{Sketch of the proposed method. The time step is fixed so that it is possible to add collision constraints for multiple quadrotors. The right bottom of the $\prescript{}{i}{\boldsymbol{\lambda}}$ is free. So that the quadrotors can 'select' freely their arrival nodes with this fixed time step strategy. The optimization target is therefore to minimize the sum of the $\prescript{}{i}{\boldsymbol{\lambda}}$ while satisfying the dynamic constraints and collision constraints, etc.} 
    \label{fig: proposed sketch}
\end{figure*}

\subsection{The optimization problem}
The aim of this paper is to generate time-optimal trajectories for a swarm of autonomous racing drones so that during the flight they can pass through required waypoints while avoiding each other and arrive at the goals with minimum time. In the CPC, a progress measure variable matrix $\boldsymbol{\lambda}$ and a progress change matrix $\boldsymbol{\mu}$ are added to their optimization problem to control the assignment of the waypoints. In particular, $\boldsymbol{\lambda}$ is used to make sure that the quadrotor can pass the pre-defined waypoints in a certain sequence and $\boldsymbol{\mu}$ is used to make sure that the distance between the assigned point on the trajectory and its corresponding waypoint is within a small tolerance. In their method, the number of nodes $N$ is determined and the quadrotor is forced to arrive at the goal at the last node. So that in the optimization process, the time step between each node $\Delta t = (t_{end}-t_{init})/N$ is decreasing with the decrease in the flight time $t_{end}$ (Fig. \ref{fig:foehn sketch}(a)). This strategy works pretty well in single racing drone scenarios. However, in terms of multiple racing drones, it is difficult to add collision constraints between each quadrotor because the number of the nodes is determined, when the quadrotors have different arrival time $\prescript{}{i}{t}_{end}$, the time of the corresponding nodes are different (Fig. \ref{fig:foehn sketch}(b)). 

To tackle the issue mentioned above, we first fix the time step $\Delta t$ and define a relatively large number of nodes $N$ so that the drones should arrive at the goal at node $\prescript{}{i}{n}_{end} < N$, which also means the arrival time $\prescript{}{i}{t}_{end} < N\Delta t$.  Fig. \ref{fig: proposed sketch}(a) shows the sketch of the scenario of two racing drones starting from the same point and flying through a waypoint and arriving at the same goal. It can be seen that since $\Delta t$ is fixed, the nodes of the quadrotors are synchronized. So that the collision constraints of the quadrotors at each node can be written as

\begin{align}
    \left\Vert \mathbf{E}(\prescript{}{i}{\mathbf{P}_k}-\prescript{}{r}{\mathbf{P}_k}) \right\Vert_2^2 - \delta_{col} \geq 0  \hspace{1cm} i\neq r
    \label{equ:constraints collision}
\end{align}

where $\prescript{}{i}{\mathbf{P}_k}$ is the position of the $i^{th}$ quadrotor at time $t_k$ and $\delta_{col}>0$ is the tolerance ensuring that two quadrotors do not collide with each other and $\mathbf{E}$ is the matrix for relieving downwash risk \cite{zhou2021ego}.

Similar to the single racing drone scenario, each quadrotor has its own $\prescript{}{i}{\boldsymbol{\lambda}}$ matrix and $\prescript{}{i}{\boldsymbol{\mu}}$ to ensure that the quadrotors fly through the waypoints in a predefined sequence and do arrive at the waypoints with the assigned node. Thus, we have the same constraints for the progress variables $\boldsymbol{\lambda}$ and progress change $\boldsymbol{\mu}$ with the CPC. 

\begin{align}
    \begin{cases}
        \prescript{}{i}{\lambda_{k}^j}  \leq   \prescript{}{i}{\lambda_{k}^{j+1}}           \\
        \prescript{}{i}{\lambda_{k+1}^j}-\prescript{}{i}{\lambda_{k}^j}+\prescript{}{i}{\mu_{k}^j} = 0 \\
        \prescript{}{i}{\mu_{k}^j}(\left \Vert \prescript{}{i}{\mathbf{P}_k} - \prescript{}{i}{\mathbf{P}}^{wj} \right \Vert_2^2-\prescript{}{i}{\nu}_k^j) := 0
    \end{cases}
    \label{equ:waypoint constraints}
\end{align}

where $\prescript{}{i}{\lambda_{k}^j}$ is a bool variable representing if the $i^{th}$ quadrotor passes the $j^{th}$ waypoint at time $t_k$. $\prescript{}{i}{\mu_{k}^j}$ means if the $i^{th}$ quadrotor is passing through the the $j^{th}$ waypoint. $\prescript{}{i}{\nu}_k^j$ is a tolerance slack. The operator '$:=$' means a NAND (not and) function.

It should be noted that in the proposed method, we don't enforce that the quadrotors arrive at the goal at the last node which is different from the CPC. It means that the quadrotors can arrive at the goal at any step if it is feasible. Fig. \ref{fig: proposed sketch}(a) shows an example of two racing drones' trajectories. It can be seen that the first quadrotor arrives at the goal at $\prescript{}{1}{t}_{end}$. The nodes after $\prescript{}{1}{t}_{end}$ will be ignored as the quadrotor has already finished its task. In this way, the optimization object is actually pushing the quadrotors' arrival node $\prescript{}{i}{n}_{end}$ leftward as further as possible while all the optimization states satisfy the constraints including the quadrotors' dynamic constraints and collision constraints, etc. Fig. \ref{fig: proposed sketch}(b) gives an example of the $1^{st}$, $w^{th}$ and $u^{th}$ ($1<w<u$) iteration of one quadrotor in the optimization process. In other words, the optimization object is minimizing the number of $\it {1}$ in the $\prescript{}{i}{\boldsymbol{\lambda}}$ matrix for all quadrotors. We define an operator $\mathbb{S}(\bullet)$ to calculate the sum of all elements in a matrix and we have the optimization target as:

\begin{align}
    \min_\mathbf{X} J = \sum_i^Q \mathbb{S}(\prescript{}{i}{\boldsymbol{\lambda}})
    \label{equ: optimation target}
\end{align}

where $Q$ is the number of quadrotors in the swarm. $\mathbf{X}$ is a set consisting of the optimization variables $\prescript{}{i}{\mathbf x}$ of all the $Q$ quadrotors, and $\prescript{}{i}{\mathbf x}=[\prescript{}{i}{\mathbf x}_0, \prescript{}{i}{\mathbf x}_1,..., \prescript{}{i}{\mathbf x}_{N-1}]$ where 
\begin{align*}
\centering
    \prescript{}{i}{\mathbf x}_k &= \begin{bmatrix}
        {\prescript{}{i}{\mathbf{p}}}_k & {\prescript{}{i}{\mathbf{v}}}_k & {\prescript{}{i}{\mathbf{q}}}_k & 
       {\prescript{}{i}{\boldsymbol{\omega}}}_k & {\prescript{}{i}{\mathbf{u}}}_k & \prescript{}{i}{\boldsymbol{\lambda}_k} & \prescript{}{i}{\boldsymbol{\mu}_k} & \prescript{}{i}{\boldsymbol{\nu}_k} 
    \end{bmatrix}
\end{align*}
are the optimization variables. By solving the optimization problem (\ref{equ: optimation target}) subject to the dynamics constraints (\ref{equ:simulation model}), collision constraints (\ref{equ:constraints collision}), waypoint constraints (\ref{equ:waypoint constraints}), inputs constraints and initial constraints, we should be able to get the optimized collision-free trajectories for each quadrotor. Compared to the CPC, in order to synchronize the nodes between quadrotors, we have to fix the time step and then estimate a relatively large node number $N$ to make sure that the quadrotors can arrive at the goals within the allowed time, which leads to the waste of the nodes after ${\prescript{}{i}{t}}_{end}$. Thus, accurately estimating the number of nodes we need is important for increasing the solving efficiency.   

\section{Simulation result and Analysis}

In this section, we analyze the optimization results and compare the proposed method with the CPC in the drone racing tracks to show its performance. And also we test the optimal trajectories in Gazebo environments to show how it works on quadrotors. We first design a racing track within a $35m \times 35m$ space as our flight arena and set $6$ waypoints within this arena. The positions of the waypoints are listed in Table \ref{tab:waypoints}.

\begin{figure}[h]
    \centering
    \includegraphics[scale=0.13, clip]{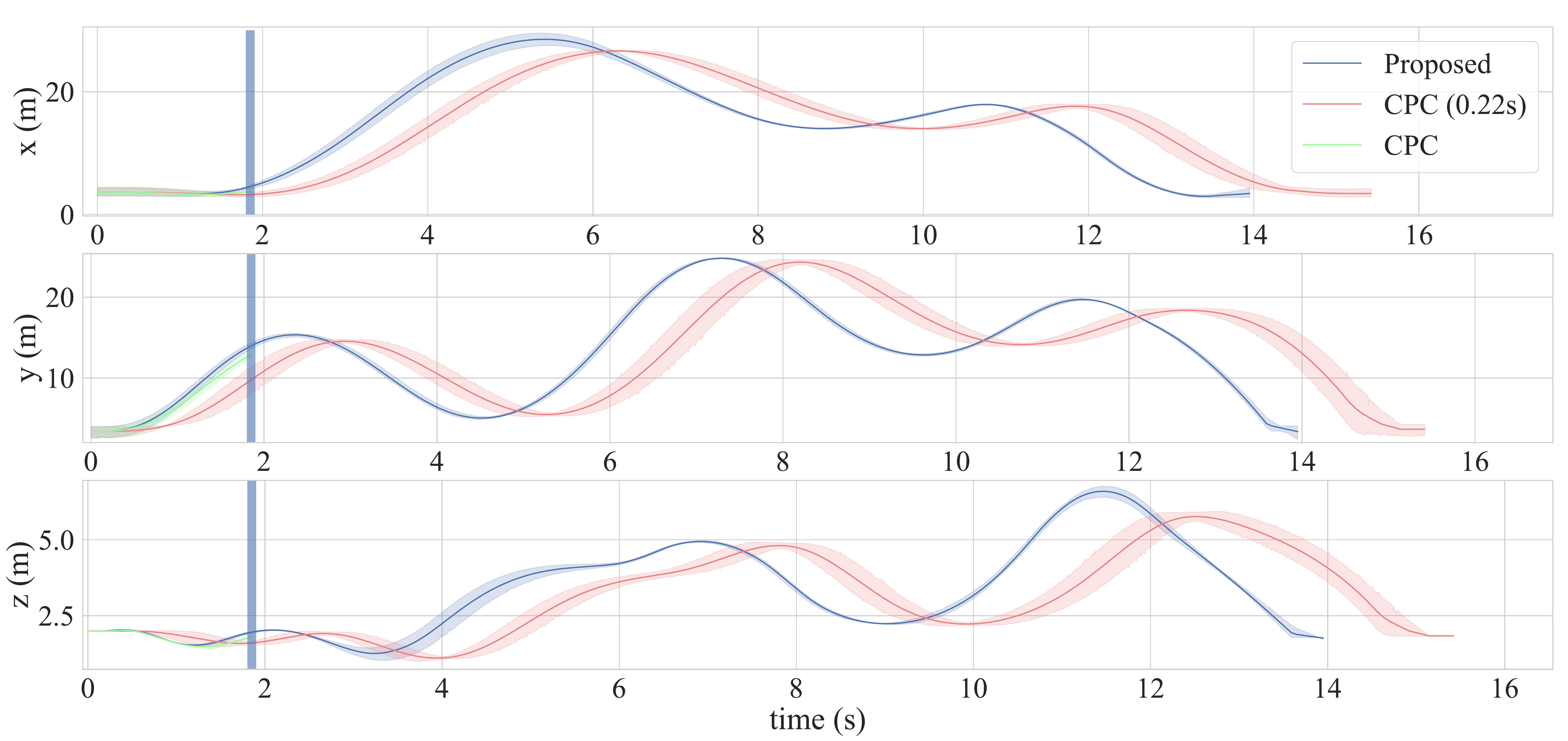}
    \caption{Comparison of the results of different trajectory generation methods for the proposed track (Table \ref{tab:waypoints}). The solid lines are the average position of the quadrotors and the shadows represent the upper and lower bounds of the quadrotors. The blue curves are the optimization results of the proposed method. The green curves are the $5$ quadrotors' trajectories which are generated independently by the benchmark method. It should be noted that  without the collision avoidance constraints, $5$ quadrotors crash at around $2s$. The red curves are similar to the green curves except that $5$ quadrotors set off with a time lag of $0.22s$.  } 
    \label{fig: optimization result}
\end{figure}

\begin{table}[h]
\caption{The position of the waypoints}
\label{table_example}
\begin{center}
\begin{tabular}{c|c|c|c}
\hline
waypoint NO. &  $x[m]$ &  $y[m]$ &  $z[m]$ \\
\hline
$1$ & $5$ & $15$ & $2$\\\hline
$2$ & $25$ & $5$ & $3$\\ \hline
$3$ & $20$ & $25$ & $5$\\ \hline
$4$ & $14$ & $14$ & $2$\\ \hline
$5$ & $18$ & $18$ & $6$\\ \hline
$6$ & $5$ & $14$ & $4$\\
\hline
\end{tabular}
\end{center}
\label{tab:waypoints}
\end{table}

We give a demonstration of $5$ racing drones flying through this track by solving the optimization problem (\ref{equ: optimation target}). Besides the waypoints' positions listed in Table \ref{tab:waypoints}, the detailed parameters are listed in Table \ref{tab:parameters}.

\begin{figure}[h]
    \centering
    \includegraphics[scale=0.3,angle=-90,trim={4cm 0cm 4cm 0cm}, clip]{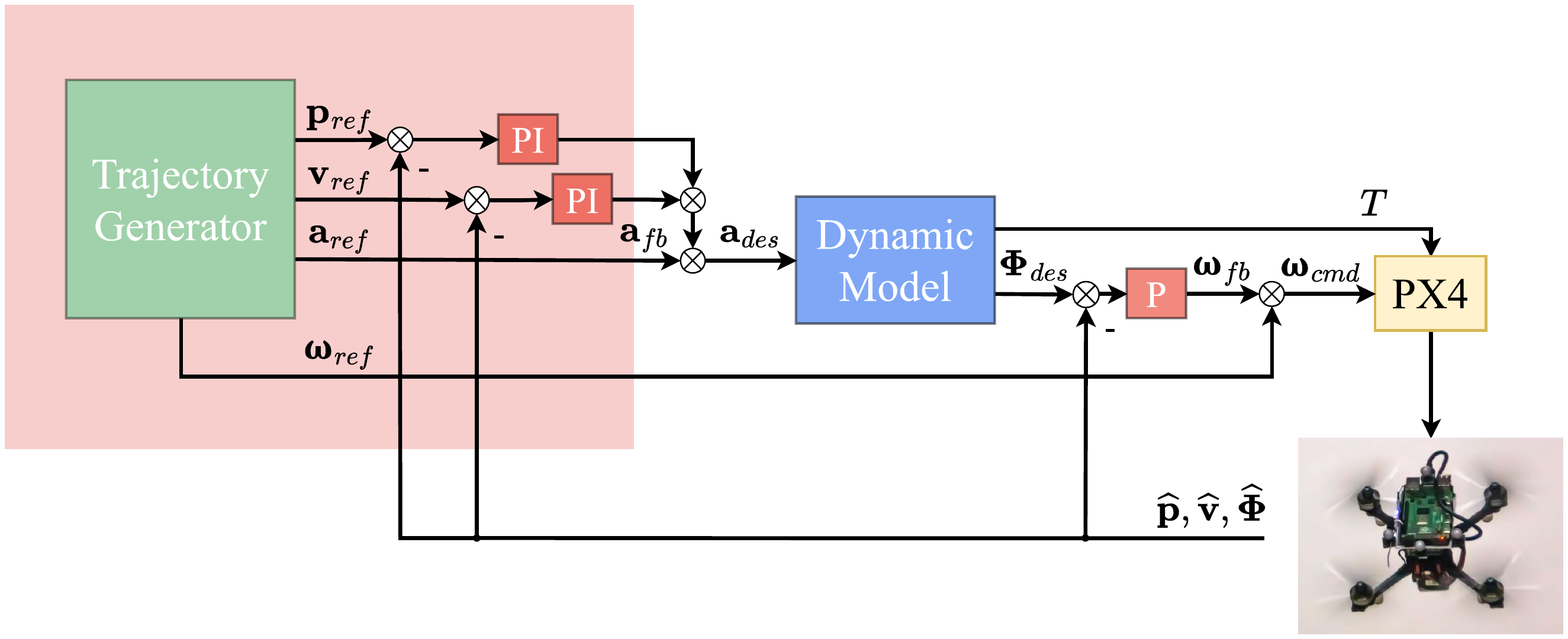}
    \caption{The control structure for the trajectory tracking. The feed-forward signal is from the generated optimal trajectories and the feedback signal is used to correct the deviation. They are implemented as a ROS node and the low-level rate loop controller is the PX4 controller. They communicate via the MAVLink protocol.} 
    \label{fig: control loop}
\end{figure}

\begin{table}[h]
\caption{The parameters used in the simulation}
\label{table_example}
\begin{center}
\begin{tabular}{|c|c|c|c|c|c|c|}
\hline
parameters & value/range & parameters & value/range \\  \hline
$N$ & $550$ & $\dot{\prescript{}{i}T_{s}}[N/s]$ & $[-120,120]$ \\  \hline
$\Delta t[s]$ & $0.03$ & $\prescript{}{i}{\theta}, \prescript{}{i}{\phi}[deg]$ & $[-60,60]$\\ \hline
$\delta_{col}[m]$ & $0.25$ & $\prescript{}{i}{\psi}[deg]$ & $[-5,5]$\\ \hline
$\prescript{}{i}{T_{s}}[N]$ & $[1,7.35]$ & $\mathbf{E}$ & $diag(1,1,1/3)$ \\ \hline 

\end{tabular}
\end{center}
\label{tab:parameters}
\end{table}

\begin{figure*}[!h]
    \centering
    \subfigure[]{\includegraphics[scale=0.06,trim={0 0cm 0cm 0}, clip]{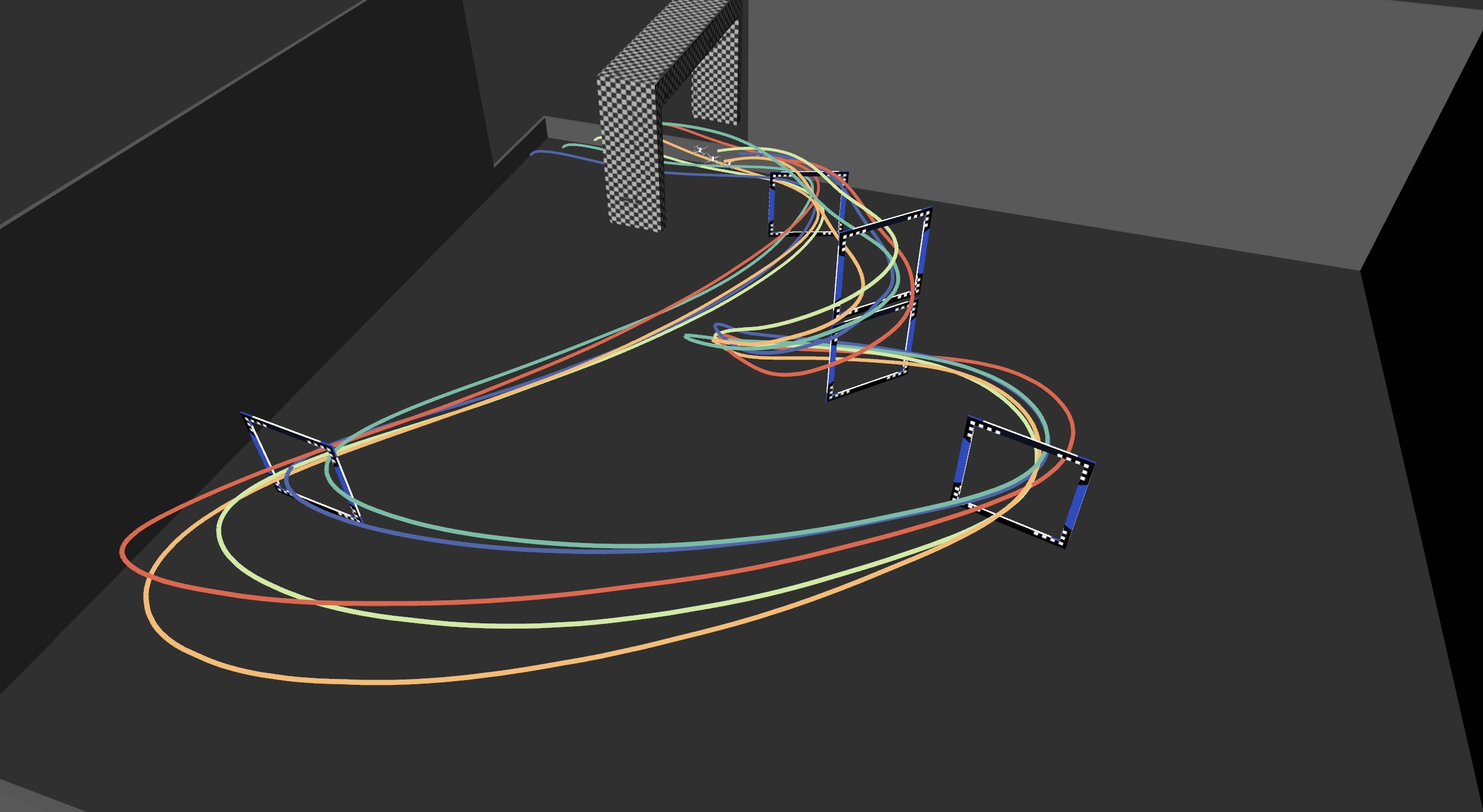}} 
    \subfigure[]{\includegraphics[scale=0.06,trim={0cm 0cm 0cm 0}, clip]{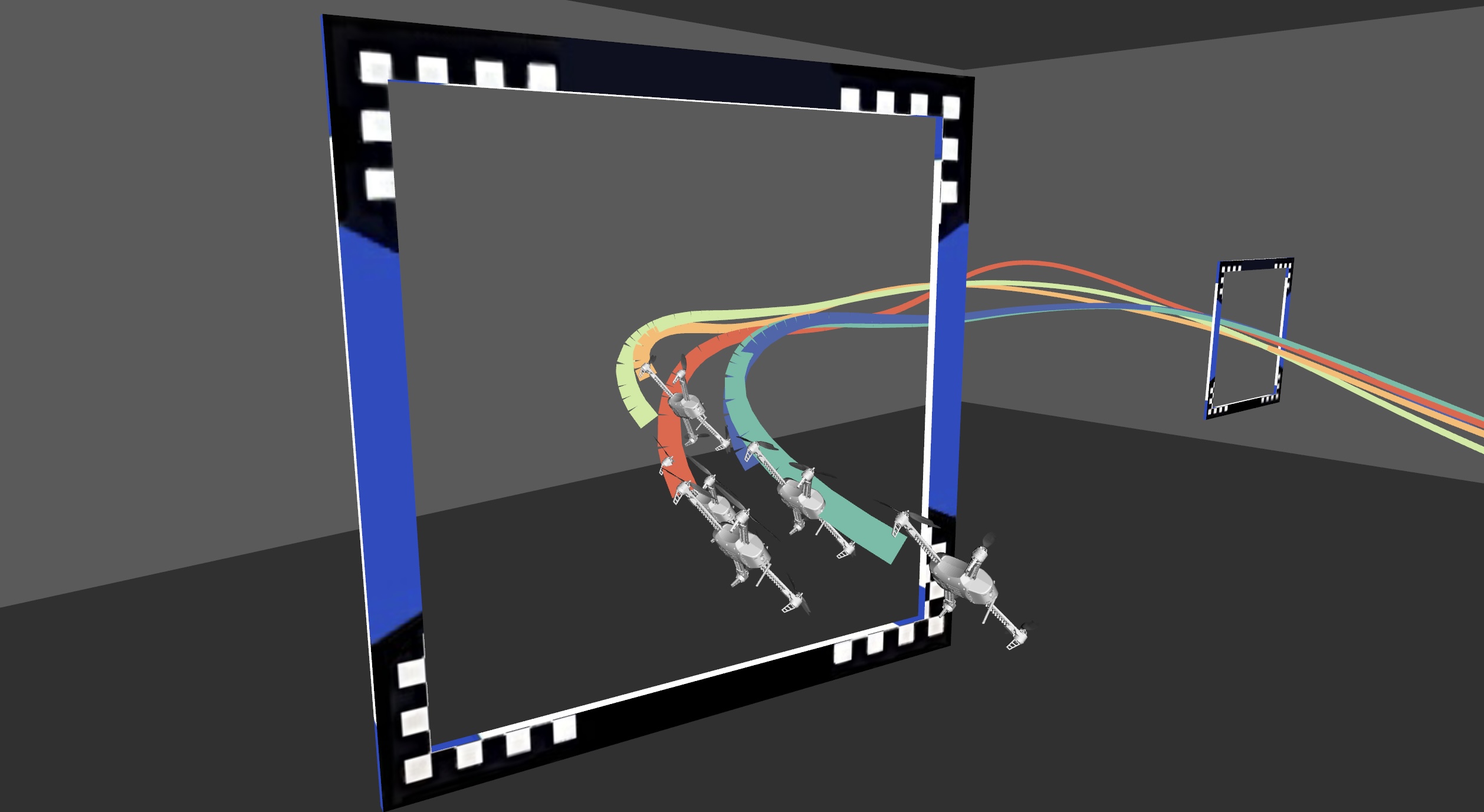}} 
        \subfigure[]{\includegraphics[scale=0.06,trim={0cm 0cm 0cm 0}, clip]{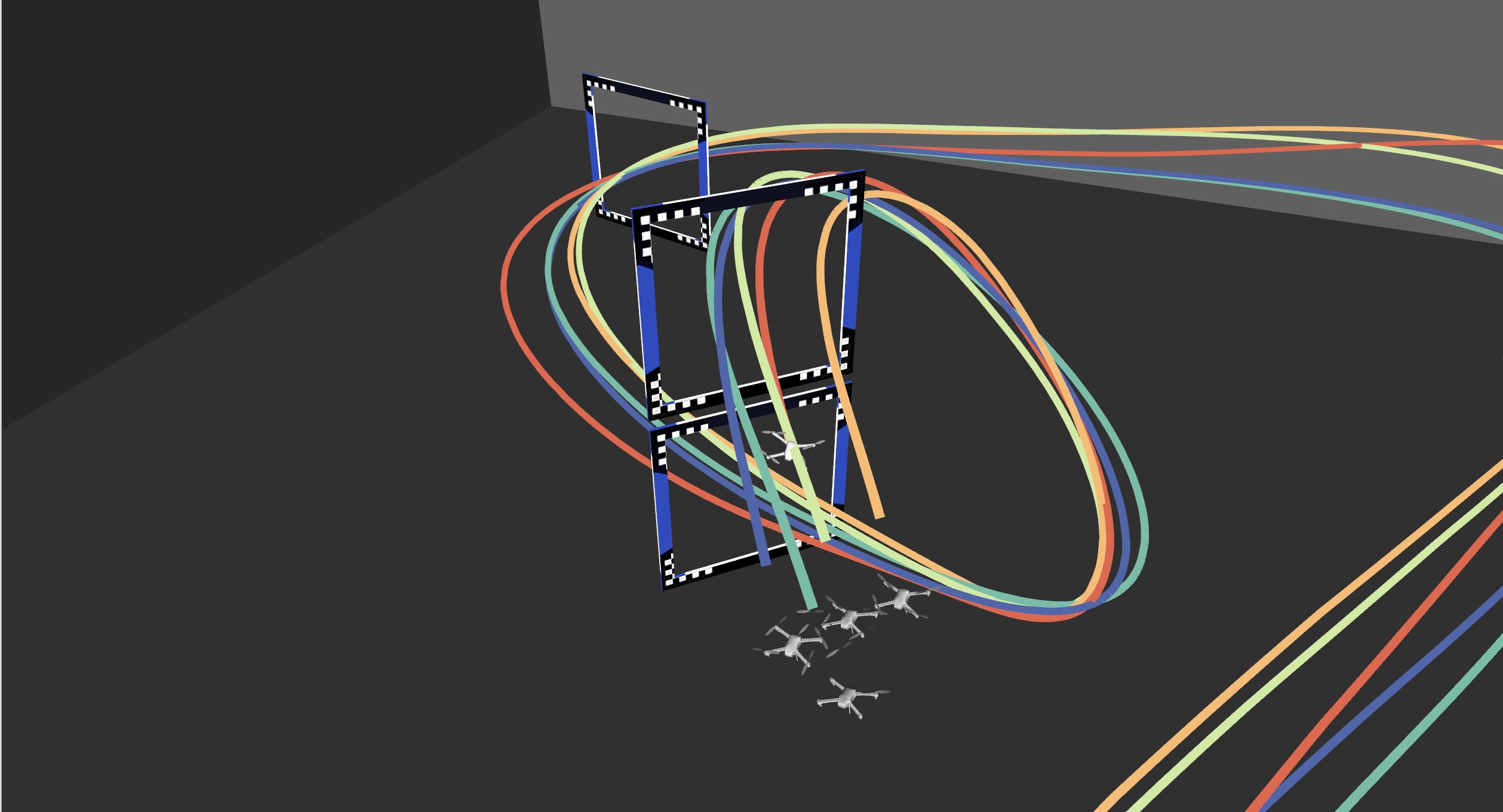}} 
    \caption{Pictures of $5$ quadrotors racing in the Gazebo simulation environment. A video is available as supplemental material.} 
    \label{fig: gazebo}
\end{figure*}

The optimization result is shown in Fig. \ref{fig: optimization result} (blue curves). To make the graphs clean, we use curves with shadows to represent the position of the swarm of quadrotors, where the solid curves are the average position of the quadrotors and the shadow boundaries are the upper and lower bounds of the quadrotors' trajectories.  To the best of the authors' knowledge, there is no research on the same topic as this paper's that multiple quadrotors race at the same time. It is difficult to find a benchmark to be compared to our proposed method. Thus, we use the CPC as the benchmark to demonstrate the performance of the proposed method that each quadrotor plans its own time-optimal trajectories independently. It is unsurprising that these quadrotors will crash at some point. The result is shown in Fig. \ref{fig: optimization result} (green curves) that they crash in $2s$ after setting off. An intuitive way to directly adopt the CPC to multi-drone racing is to have the quadrotors set off with some time lag. We find that a time lag of $0.22s$ can ensure these quadrotors don't collide with each other and the result is shown in Fig. \ref{fig: optimization result} (red curves). It can be seen that the proposed method and the benchmark method with $0.22$ time lag can guide the quadrotors to the goal but the total arrival time of the proposed method is $0.9s$ less than the benchmark method. It should be noted that we artificially add time lags for the quadrotor which can help them avoid collisions. However, this method cannot guarantee collision-free in all scenarios like a simple back-and-forth trajectory.      

We then test the trajectories in the Gazebo simulator to demonstrate the performance of the trajectory tracking. The controller we use for low-level control is the commonly used open-source PX4. The high-level controller is written as a ROS node that communicates with the PX4 via ROS topics. The controller is a classic feed-forward and feedback strategy that the generated time-optimal rotation rates and thrust serve as feed-forward signals and a classic two-loop PID controller serves as the feedback controller to correct the deviation (Fig. \ref{fig: control loop}).

\begin{figure}[h]
    \centering
     \subfigure[$5$ quadrotors set off together and fly through the waypoints without collision]{\includegraphics[scale=0.18,trim={10cm 0cm 10cm 0cm},clip]{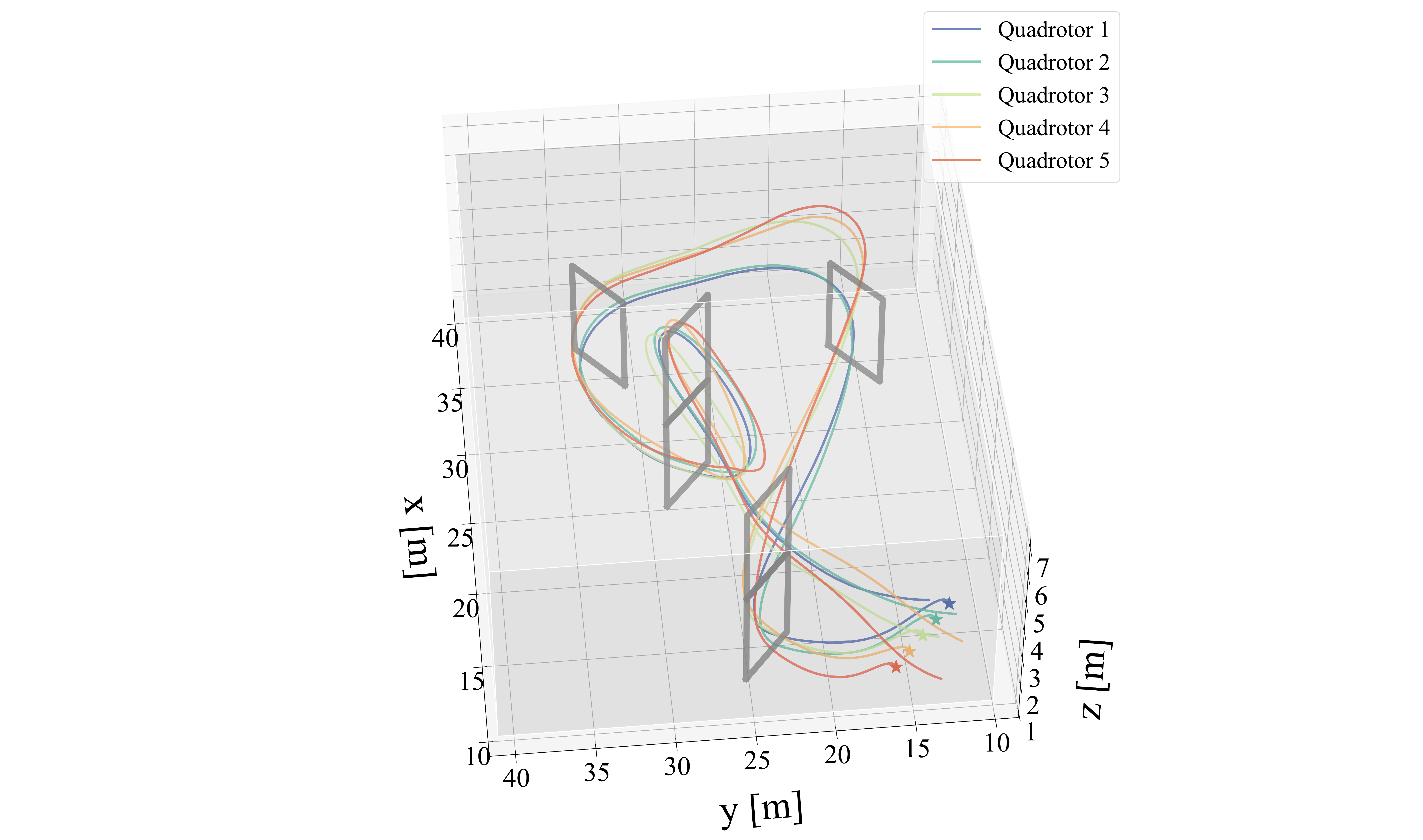}} 
     \subfigure[The speed distribution of the quadrotors. The maximum speed achieves $14m/s$]{\includegraphics[scale=0.12,trim={0cm 0cm 0cm 0cm }]{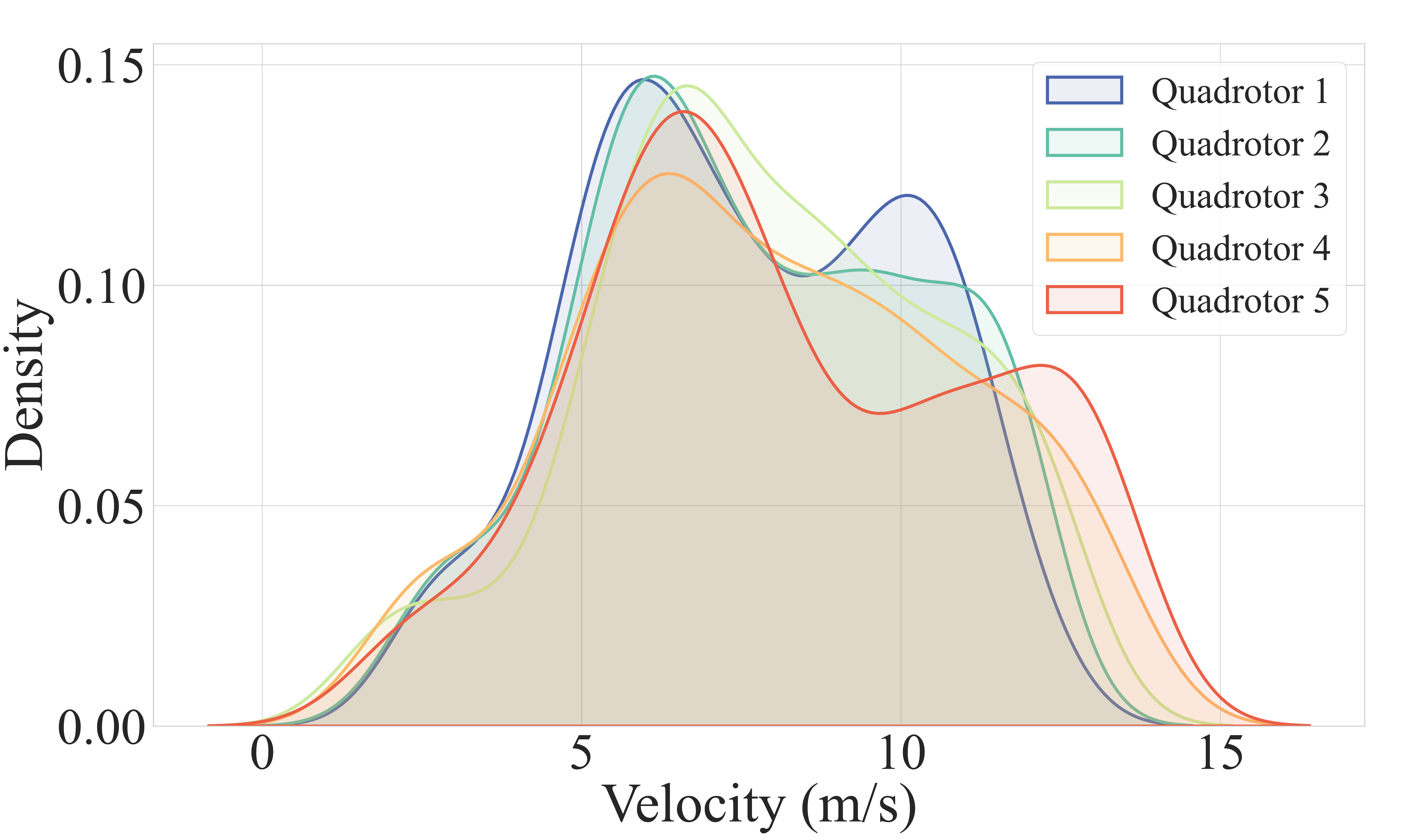}} 
    \caption{Flight data of the simulation.}
    \label{fig:simulation result} 
\end{figure}

The simulation scenarios in Gazebo are shown in Fig. \ref{fig: gazebo}. The flying arena is a $35m \times 35m$ space and the waypoints (racing gates) are deployed at the positions listed in Table \ref{tab:waypoints}. $5$ quadrotors set off at the same time and fly through the gates according to the pre-defined sequence without collision. Fig. \ref{fig:simulation result} shows the trajectories of the $5$ quadrotors and their speed distribution. It can be seen the maximum flying speed in this scenario achieves $14m/s$. 

\section{experiment setup and result}

In this section, we show the flight performance of the proposed method in the real world. The flying platform is a self-made quadrotor with a Raspberry Pi onboard running Ubuntu 20.04 and ROS Noetic to run the high-level controller (Fig. \ref{fig:two drone racing photo}). A CUAV nora+ autopilot running PX4 is used for low-level control. The high-level controller sends angular rate and throttle commands to the autopilot via MAVLink protocol. The autopilot executes the commands to control the quadrotor to track the planned trajectory. The flying platform weighs $713g$ and has a thrust-to-weight ratio of $5$. The Opti-track system is used in the experiment to provide accurate position feedback. 

\begin{figure}
    \centering
    \includegraphics[scale=0.17, clip]{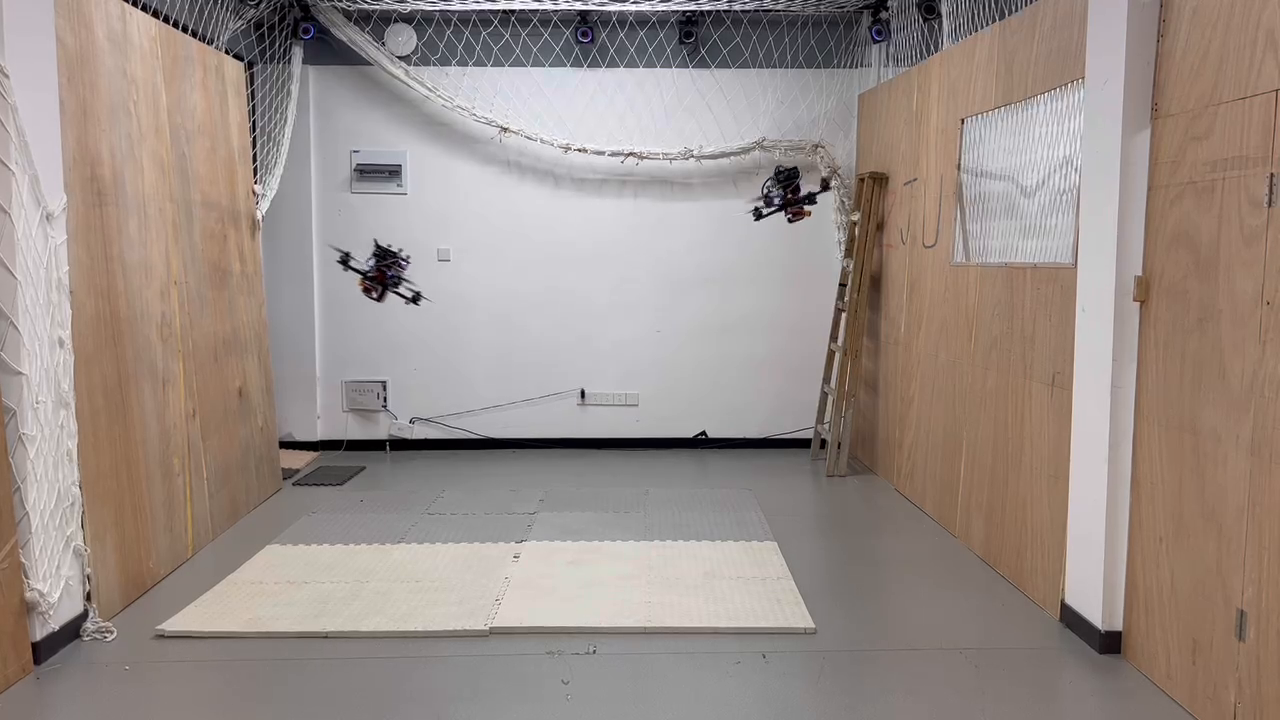}
    \caption{Picture of two quadrotors flying in the arena.} 
    \label{fig: flying platform}
\end{figure}

\begin{figure}[h]
    \centering
    \subfigure[The collision-free trajectories of two quadrotors in the real-world experiment.]{\includegraphics[scale=0.15,trim={15cm 0cm 12cm 0cm}, clip]{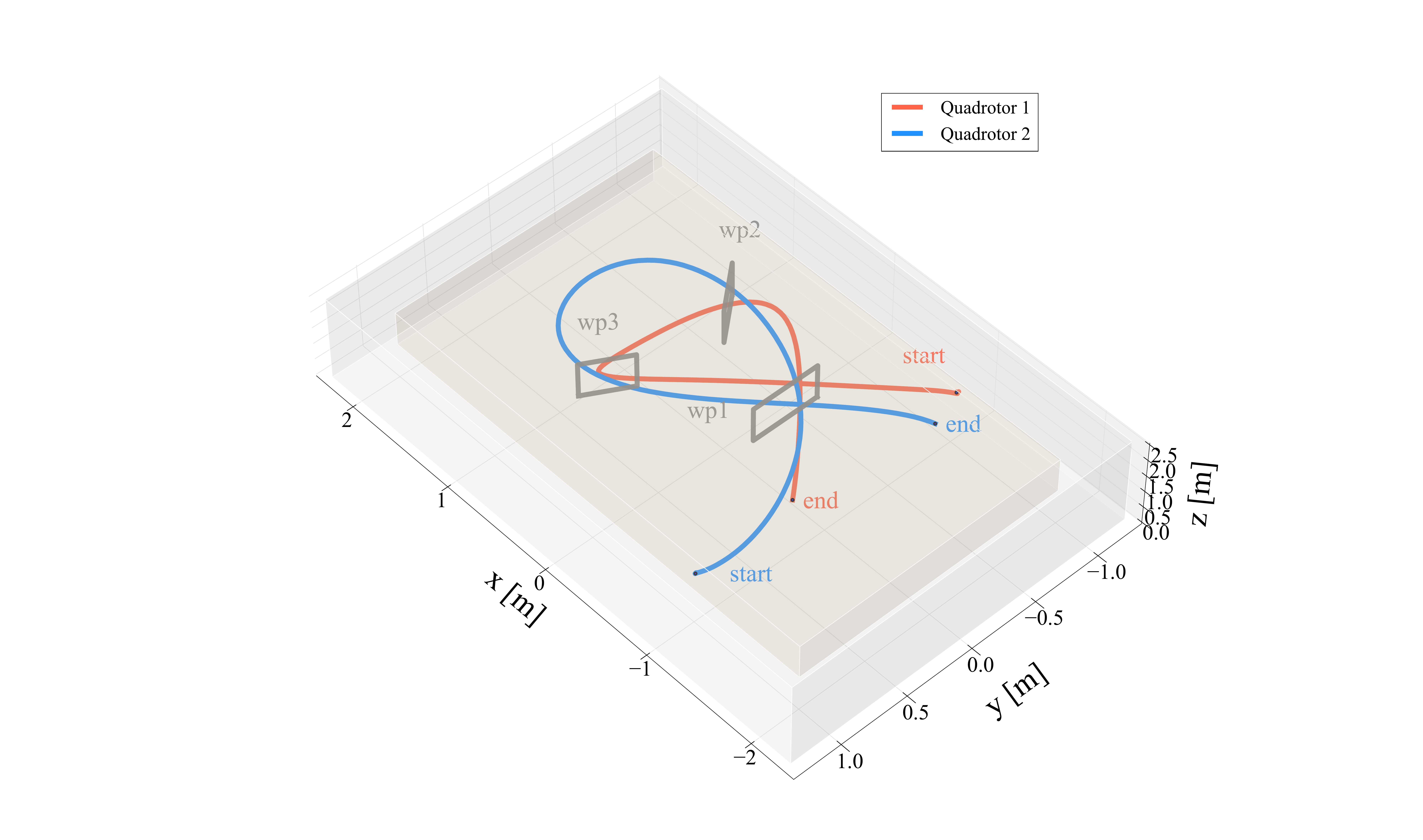}} 
        \subfigure[The speed distribution of the quadrotors. The maximum speed is $2m/s$]{\includegraphics[scale=0.125 ,trim={0cm 0cm 0cm 0}, clip]{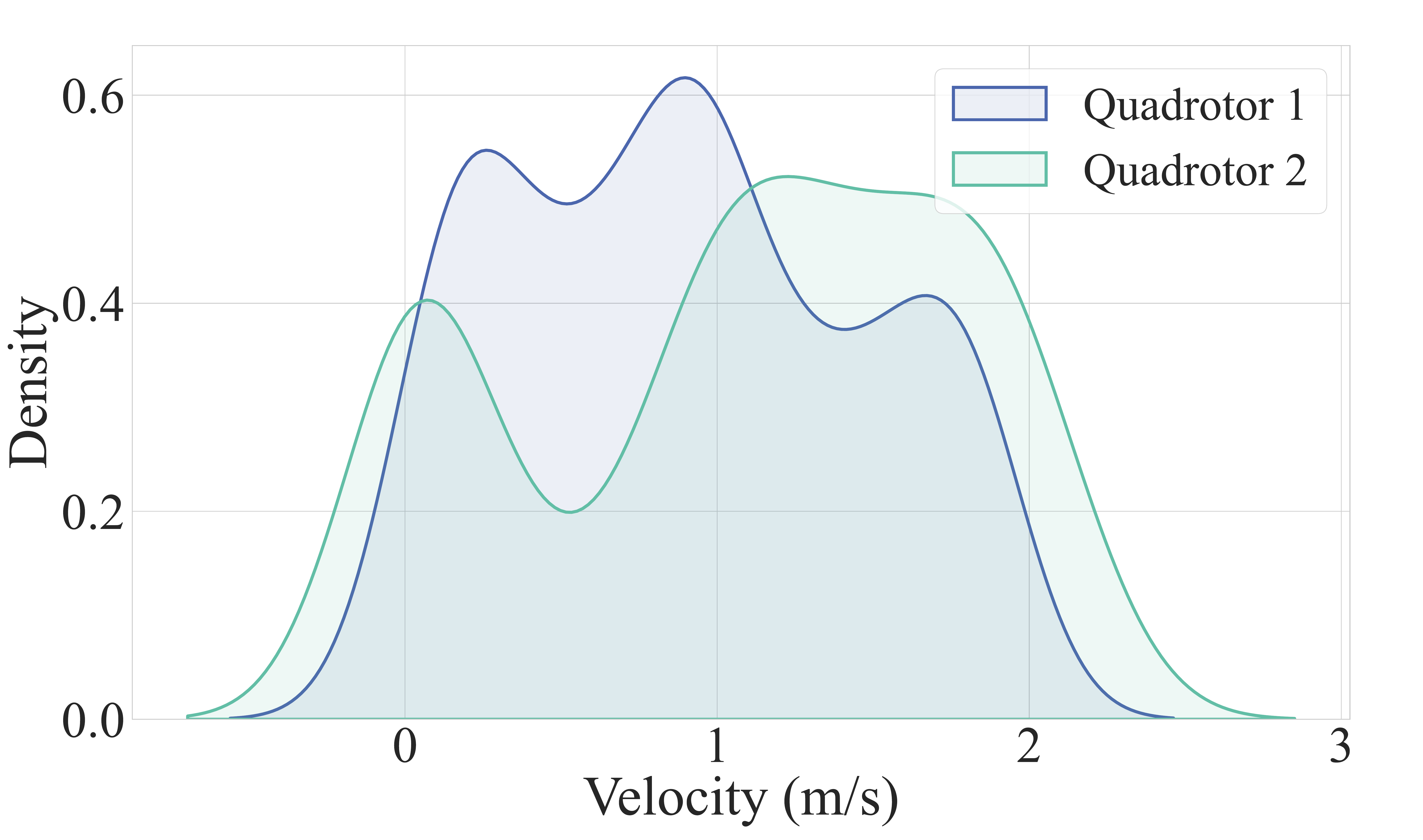}} 
    \caption{The flying result of two quadrotors flying a $3$-waypoint track.} 
    \label{fig: real world data}
\end{figure}


Due to the limited size of the Opti-track system's coverage ($4.6m \times 2.6m \times 2.5m$) and the necessary safety margin, the free-flying space is only $4m \times 2m \times 1m$. Thus, we only have $3$ waypoints to generate trajectories to show how the proposed method performs in the real world. The quadrotors' waypoints are listed in Table \ref{tab: real-world waypoints}.

\begin{table}[h]
\setlength{\belowdisplayskip}{0pt}
\caption{The position of the waypoints in the real-world experiment}
\label{table_example}
\begin{center}
\begin{tabular}{c|c|c|c}
\hline
quadrotor ID  & waypoint 1   & waypoint 2 & waypoint 3 \\ \hline
$1$ & $(0,0,1.5)$ & $(1,-1,1.5)$ & $(1,1,1.5)$  \\ \hline
$2$ & $(0,0,1.5)$ & $(1,1,1.5)$ & $(1,-1,1.5)$  \\ \hline
\end{tabular}
\end{center}
\label{tab: real-world waypoints}
\end{table}
\setlength{\tabcolsep}{3pt}

In the real-world experiment, the quadrotors have to accelerate and decelerate in the extremely limited maneuver space, the highest flying speed is $2m/s$. The trajectories and speed distribution are shown in Fig. \ref{fig: real world data} where we use the gates to represent the waypoints.  It can be seen that the two quadrotors can fly through the required waypoints while avoiding each other and arrive at their goals with minimum time.



\section{CONCLUSIONS}

In this paper, we propose a novel autonomous multi-drone racing trajectory generation approach that generates the time-optimal trajectories passing waypoints in sequence while avoiding collisions. We also demonstrate in simulation that the proposed method can generate trajectories for $5$ quadrotors with $6$ waypoints in a $35m \times 35m$ space. In this challenging environment, the quadrotors can achieve a maximum speed of $14m/s$. We also test the proposed method in real-world experiments, due to the extremely limited size of the flying arena, the quadrotors can fly through the waypoints without collision with a top speed of $2m/s$, which demonstrates the feasibility of our proposed method in the real world.

There are also many research directions to explore in the future. For example, how to improve the optimization efficiency or even move the optimization onboard the quadrotor is worthwhile to discuss. Also how to improve the trajectory tracking performance for aggressive maneuvers is another interesting topic to study. Additionally, for micro aerial vehicles, localization with high speed using fully onboard resources is still a big challenge that needs to be solved to make the robots fly outside the laboratories.  




\bibliography{root}
\bibliographystyle{ieeetr}

\end{document}